\documentclass[letterpaper, 10 pt, journal, twoside]{IEEEtran}

\IEEEoverridecommandlockouts                              

\usepackage[english]{babel}

\usepackage{graphicx}
\usepackage{animate}
\usepackage{epsfig} 				
\usepackage{epstopdf}

\usepackage{listings}
\usepackage{color}
\usepackage{nameref}

\usepackage{xcolor}
\usepackage[colorlinks,bookmarks=false,citecolor=blue,linkcolor=blue,urlcolor=black]{hyperref}
\usepackage[numbers]{natbib}	 			

\usepackage{amsmath}	 			
\usepackage{amssymb}  				

\usepackage{dsfont}			
\usepackage{mathtools}

\usepackage{epigraph}
\usepackage{lscape}
\usepackage[]{nomencl}				
\usepackage{algorithm}
\usepackage{algorithmic}
\usepackage{multicol}
\usepackage{multirow}
\usepackage{makecell}
\usepackage{etoolbox}
\usepackage{graphics}

\usepackage{wrapfig}

\usepackage{siunitx}

\usepackage{floatflt}

\usepackage{url}

\usepackage{placeins}
\usepackage{bm}

\usepackage{booktabs}
\usepackage{textcomp,mathcomp}
\usepackage{bm}

\newcommand{\email}[1]{\href{mailto:#1}{#1}}

\newcommand{\link}[1]{\colora{\url{#1}}}

\renewcommand{\sec}[1]{Section~\ref{#1}}
\newcommand{\fig}[1]{Fig.~\ref{#1}}

\newcommand{\tab}[1]{Table~\ref{#1}}

\newcommand{\ProjectWeb}[0]{\href{https://linchangyi1.github.io/9DTact}{https://linchangyi1.github.io/9DTact}}

\newcommand{\sensor}[0]{9DTact} 

\newcommand{\revision}[1]{#1}
\newcommand{\cd}[1]{#1}

\title{\LARGE \bf \sensor: A Compact Vision-Based Tactile Sensor for Accurate 3D Shape Reconstruction and Generalizable 6D Force Estimation}

\author{Changyi Lin, Han Zhang, Jikai Xu, Lei Wu and Huazhe Xu
\thanks{Manuscript received: July, 24, 2023; Revised October, 11, 2023; Accepted November, 25, 2023. This paper was recommended for publication by Editor A. Banerjee upon evaluation of the Associate Editor and Reviewers' comments. (\textit{Corresponding author: Huazhe Xu})}
\thanks{Changyi Lin is with Shanghai Qi Zhi Institute, Shanghai 200030, China, and also with Institute for Interdisciplinary Information Sciences, Tsinghua University, Beijing 100084, China. (e-mail: \email{changyil@andrew.cmu.edu})}
\thanks{Han Zhang is with Department of Electronic Engineering, Tsinghua University, Beijing 100084, China, and also with Shanghai Qi Zhi Institute, Shanghai 200030, China. (e-mail: \email{zhanghan14@mails.tsinghua.edu.cn})}
\thanks{Jikai Xu and Lei Wu are with School of Mechanical Science and Engineering, Huazhong University of Science and Technology, Wuhan 430074, China, and also with Shanghai Qi Zhi Institute, Shanghai 200030, China. (e-mail: \email{jikai\_xu@hust.edu.cn}; \email{lei\_wu@hust.edu.cn})}
\thanks{Huazhe Xu is with the Institute for Interdisciplinary Information Sciences, Tsinghua University, Beijing 100084, China, and also with Shanghai Qi Zhi Institute, Shanghai 200030, China, as well as with Shanghai Artificial Intelligence Laboratory, Shanghai 200032, China. (e-mail: \email{huazhe\_xu@mail.tsinghua.edu.cn})}
\thanks{Digital Object Identifier (DOI): see top of this page.}
}

\usepackage[font=footnotesize]{caption}

\begin{document}

\maketitle


\begin{abstract}
	The advancements in vision-based tactile sensors have boosted the aptitude of robots to perform contact-rich manipulation,
particularly when precise positioning and contact state of the manipulated objects are crucial for successful execution.
In this work, we present \underline{9D}Tact, a straightforward yet versatile tactile sensor that offers \underline{3D} shape reconstruction and \underline{6D} force estimation capabilities.
Conceptually, \sensor{} is designed to be highly compact, robust, and adaptable to various robotic platforms. Moreover, it is low-cost and \revision{easy-to-fabricate}, requiring minimal assembly skills.
Functionally, \sensor{} builds upon the optical principles of DTact and is optimized to achieve 3D shape reconstruction with enhanced accuracy and efficiency.
Remarkably, we leverage the optical and deformable properties of the translucent gel so that \sensor{} can perform 6D force estimation without the participation of auxiliary markers or patterns on the gel surface. 
More specifically, we collect a dataset consisting of approximately 100,000 image-force pairs from 175 complex objects and train a neural network to regress the 6D force, which can generalize to unseen objects.
To promote the development and applications of vision-based tactile sensors, we open-source both the hardware and software of \sensor{}, along with a comprehensive video tutorial, all of which are available at \ProjectWeb.

\end{abstract}

\begin{IEEEkeywords}
Force and Tactile Sensing; Perception for Grasping and Manipulation
\end{IEEEkeywords}


\section{Introduction}

\IEEEPARstart{T}{actile} sensing, which provides physical properties and spatial state of contact objects, is vital for robots to interact with the real world. With the help of cameras, 
vision-based tactile sensors~\cite{lin2023dtact, yuan2017gelsight, taylor2022gelslim, lambeta2020digit, gelsight-mini2022} are able to sense the deformation of the gel surface approaching human-scale resolution. The acquired high-resolution information enables robots to perform stable and accurate robotic manipulation tasks such as object insertion~\cite{li2014localization, dong2021tactile} and cable manipulation~\cite{she2021cable, wilson2023cable}. Despite the potential benefits of these sensors, their adoption within the robotics community has been limited due to various factors such as the lack of compactness, the complexity of fabrication, high cost of acquisition, fragility and instability during use, and deficient functional capabilities.

In this work, we present \sensor{}, a vision-based tactile sensor equipped with the following merits, aiming to overcome the drawbacks of previous counterparts.
\begin{itemize}
    \item {\textbf{Hardware.} Our sensor not only excels in specifications, it can also be fabricated in a convenient manner.
    With iterations of optimization in illumination, structure, crafts, and materials, \sensor{} is designed to be compact, robust, and adaptable to various robotic platforms. Furthermore, \sensor{} is an affordable and easily assembled sensor that only requires accessible components, standard machining processes, and minimal assembly skills.}
    \item {\textbf{Software.} \sensor{} is a versatile sensor capable of both accurate 3D shape reconstruction and generalizable 6D (1D normal, 2D shear, and 3D torque) force estimation.
    The new design and simple calibration method improve the accuracy and efficiency of 3D shape reconstruction.
    Inspired by the principles of DTact~\cite{lin2023dtact} that pixels corresponding to thinner areas become darker, we observe another interesting phenomenon that pixels corresponding to bulging areas become brighter, and the in-plane motion of the contact object induces accompanying movement of the brighter pixels.
    Based on this finding, we extract a dense deformation representation from the original tactile image for force estimation, where no auxiliary markers or patterns are needed. Making use of a neural network trained on approximately 100,000 pairs of deformation representation and 6D force sampled from 175 objects, \sensor{} could estimate accurate 6D force with generalization to unseen geometries and objects.}
    \item {\textbf{Open-Source.}
    We would like to clear the obstacles as possible for building and utilizing tactile sensors such as \sensor{} in the robotics community.
    Hence, we open-source everything about \sensor{} including its design files, codes, datasets, and pre-trained models. Furthermore, we also provide a \revision{comprehensive} video tutorial that documents the entire process of replicating a \sensor{} sensor, including a bunch of experiences for simplifying and improving the manufacturing processes.}
\end{itemize} 

The remainder of this paper is organized as follows. We introduce related work on hardware designs and force estimation methods of vision-based tactile sensors in~\sec{sec:related}. The details of \sensor{} design are described in~\sec{sec:design}. We then introduce the improvements of 3D shape reconstruction in~\sec{sec:shape_reconstruction}. Next, we present the principle, implementation, and performance of 6D force estimation in~\sec{sec:force_estimation}. Finally, the conclusion is summarized in~\sec{sec:conclusion}.
 %
	

\section{Related Work}
\label{sec:related}

	\begin{table*}[ht]
\centering
\caption{Comparison of GelSight, GelSlim 3.0, DIGIT, GelSight-Mini, DTact, and \sensor{}. ($^*$Manufacturing of 1000 pieces. $^\dagger$Commodity price.)}
\begin{tabular}{lccccccc}\toprule
Sensor & Dimension [$mm^3$] $\bm{\downarrow}$& Sensing Area [$mm^2$] $\bm{\uparrow}$& D/A Ratio [$mm$] $\bm{\downarrow}$& Weight [$g$] $\bm{\downarrow}$&FPS $\bm{\uparrow}$& Cost[\$] $\bm{\downarrow}$\\
\midrule
GelSight~\cite{dong2017improved}& $80\times40\times40 = 128000$& 252 & 508 &NA&\textbf{90}& 30 \\
GelSlim 3.0~\cite{taylor2022gelslim}& $80\times37\times20 = 59200$&$\textbf{675}$& 88 &$45$&\textbf{90}&$25^*$\\
DIGIT~\cite{lambeta2020digit}&$36\times26\times33 = 30888$&$19\times16=304$& 102 & $\textbf{20}$&60&$15^*$ / $300^\dagger$\\
GelSight-Mini~\cite{gelsight-mini2022}&$32\times28.5\times28 = 25536$&$19\times15 = 285$& 90 & 20.8 & 25 &$499^\dagger$\\
DTact~\cite{lin2023dtact}&$45\times45\times47 = 95175$&$24\times24=576$& 165 & 78 & 60 & 34 \\
\midrule
\textbf{\sensor{} (Ours)}&$32.5\times25.5\times25.5 = \textbf{21133}$&$24\times18=432$& \textbf{49} & \textbf{20} & \textbf{90} & \textbf{15}\\
\bottomrule
\end{tabular}
\label{tab:sensors-comparison}
\end{table*}

\begin{table*}[ht]
\centering
\caption{Comparison of the methods and configurations utilized for force estimation by GelSlim 2.0, DelTact, DenseTact 2.0, and \sensor{}.}
\begin{tabular}{lccccc}\toprule
Sensor & Deformation Visualizer & Deformation Representation & Infer Method & Validation Objects & Collection State\\
\midrule
\revision{GelSight~\cite{yuan2017gelsight}} & \revision{Black marker array} & \revision{Marker pixels in raw image} & \revision{CNNs~\cite{lecun1998gradient}} & \revision{3 simple objects} & \revision{Dynamic} \\
GelSlim 2.0~\cite{ma2019dense} & Black marker array & 3D motions of markers & iFEM~\cite{bathe2006finite} & 1 sphere & Static \\
DelTact~\cite{zhang2022deltact, zhang2019effective} & Colorful dense pattern & Vectors field from optical flow & NHHD~\cite{bhatia2014natural} & 5 spheres & Static \\
DenseTact 2.0~\cite{do2023densetact} & Black randomized pattern & Pattern pixels in raw image  & CNNs & 10 \revision{simple} objects & Static \\
\midrule
\textbf{\sensor{} (ours)} & Only original gel & Dense gel flow image& CNNs & 175 \revision{complex} objects & Dynamic \\
\bottomrule
\end{tabular}
\label{tab:Force-comparision}
\end{table*}

\subsection{Compact Vision-Based Tactile Sensors for 3D Shape Reconstruction}
\label{sec:related_sensors}
Vision-based tactile sensor GelSight~\cite{yuan2017gelsight} leverages the photometric stereo technique~\cite{johnson2009retrographic} to achieve 3D shape reconstruction of its sensing surface. Following, many GelSight-like sensors~\cite{dong2017improved, wang2021gelsight, taylor2022gelslim} improve the designs to be more compact to mount them in grippers or dexterous hands. However, their reliance on the photometric stereo technique, which is highly dependent on the uniformity and reflection of the internal illumination, makes them challenging to replicate due to strict requirements on material preparation, fabrication processes, and assembly skills. Consequently, researchers lacking hardware experience must purchase commercial products~\cite{lambeta2020digit, gelsight-mini2022}.

To address the manufacturing challenges mentioned above, DTact~\cite{lin2023dtact} \revision{leverages the reflection property of translucent elastomer} for 3D shape reconstruction,  which has demonstrated comparable accuracy, superior robustness and surface shape extensibility. Based on this promising principle, \sensor{} is carefully designed to simultaneously possess exceptional physical characteristics including compactness, robustness, and affordability as highlighted in~\tab{tab:sensors-comparison}. Moreover, \sensor{} is \revision{easy-to-fabricate} and open-sourced, which expands its potential as a general vision-based tactile sensor.

\subsection{Deformation Representation for 6D Force Estimation}
\label{sec:related_force}
On vision-based tactile sensors with gel surfaces, an applied force induces deformation. Therefore, the methods for force estimation generally consist of the following three key elements.
\begin{itemize}
    \item {\textbf{Deformation visualizer:} the physical medium for visualizing the full-dimensional deformation of the gel into visual features for the camera.}
    \item {\textbf{Deformation representation:} the information extracted from the visual features.}
    \item {\textbf{Inferring method:} the method for decoupling force from the deformation representation.}
\end{itemize}

\revision{As outlined in~\tab{tab:Force-comparision}, tactile sensors capable of 3D shape reconstruction often incorporate a marker array or a pattern on their sensing surface as a common method for force estimation. GelSight~\cite{yuan2017gelsight} employs Convolutional Neural Networks~(CNNs)~\cite{lecun1998gradient} to predict force from tactile images, while GelSlim~\cite{ma2019dense} utilizes the inverse Finite Element Method~(iFEM)~\cite{bathe2006finite} to infer force based on the 3D motions of markers. However, the sparsity of the marker array limits its capacity to fully capture deformation across the entire gel surface, leading to validations on only a few selected objects with simple geometries.}
Similar deformation representation is used in DelTact~\cite{zhang2022deltact} but with more data for computing the coefficient matrix. To enrich the information contained in the deformation representation, DenseTact 2.0~\cite{do2023densetact} develops a special fabrication process to paint a continuous pattern, and employs CNNs for predicting 6D force.

\revision{Although 9DTact could potentially adopt the method of painting pattern for force estimation, there are inherent drawbacks with this method. Specifically, the painted pattern, being non-reflective to directional lights, compromises the sensor's 3D shape reconstruction capability within the regions it covers~\cite{wang2021gelsight}, meaning that there is a trade-off between the two functionalities. Moreover, crafting the pattern necessitates specialized equipment and expertise. Consequently, the method of painting pattern is incompatible with our objectives of maintaining the accuracy of 3D shape reconstruction and simplifying the sensor fabrication.}

Surprisingly, we find that \sensor{}, without any painted pattern, naturally possesses a dense gel flow. Such flow is innate from the optical and deformable properties of the translucent gel. Moreover, the flow reflects the full-dimensional deformation of the gel. This phenomenon not only helps extract dense deformation representations to perform 6D force estimation with generalization to unseen objects, but also preserves the 3D shape reconstruction quality, \revision{and simplifies} the fabrication.


\section{\sensor\ Sensor Design} 
\label{sec:design}

	\subsection{Design Goals}
Throughout the fabrication, installation, and utilization of the sensor, we aim to achieve the following objectives.
To lower the barrier of fabrication, the components should be readily accessible and the assembly process should require minimal professional expertise.
For installation, the sensor should possess a compact structural configuration that allows installation in \revision{constrained} spaces, such as within \revision{the fingertips of} dexterous hands.
Regarding utilization, the sensor should exhibit robustness in both physical and functional aspects, as well as adaptability to various computing platforms.

\begin{figure}[t]
	\centering
	\includegraphics[width= \linewidth]{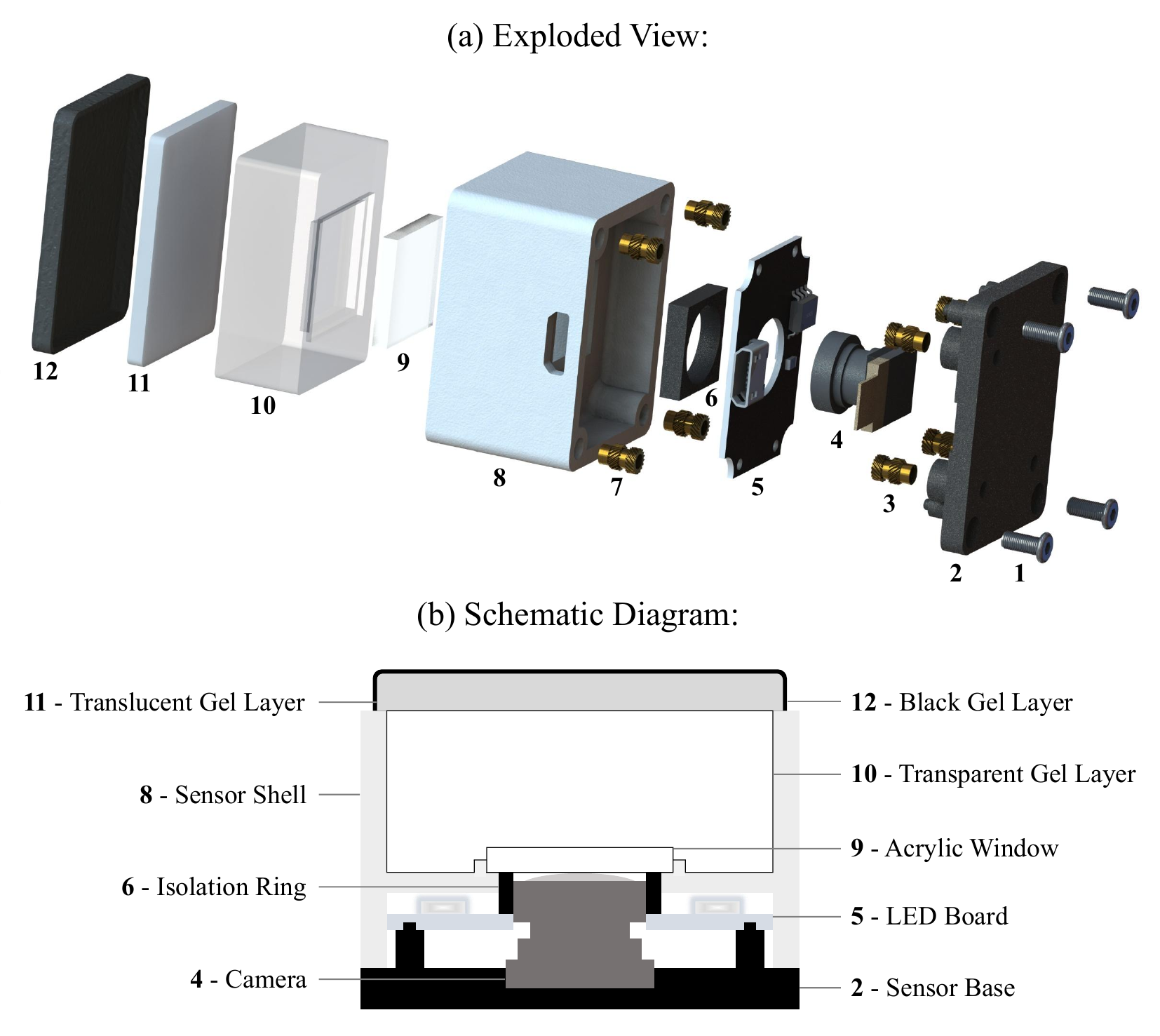}
	\caption{Design of \sensor{}. (a) The exploded view of \sensor{}. The components labeled as 1 are screws for connecting the sensor base and the sensor shell. The components labeled as 3 and 7 are both heat-set threaded inserts. (b) The schematic diagram of \sensor{}.}
	\label{fig:design}
\end{figure}

In the following section, we will demonstrate how we achieve these goals by providing a detailed description of each component shown in~\fig{fig:design} (b).

\subsection{Details of the Components}
\label{sec:design_details}
\textbf{Camera.} In order to capture the sensor's contact surface as comprehensively as possible, we select an OV5647 camera with a wide Field Of View~(FOV) of 160 degrees and attach it to the sensor base using 3M glue. The camera occupies a small space and is also adaptable to various computing platforms. It can be connected to the Camera Serial Interface~(CSI) port of Raspberry Pi Zero directly or to the Universal Serial Bus~(USB) port with an off-the-shelf CSI-to-USB transformation board. In this work, we choose the latter connection format to use the camera with a desktop.

\begin{figure*}[t]
	\centering
	\includegraphics[width= \linewidth]{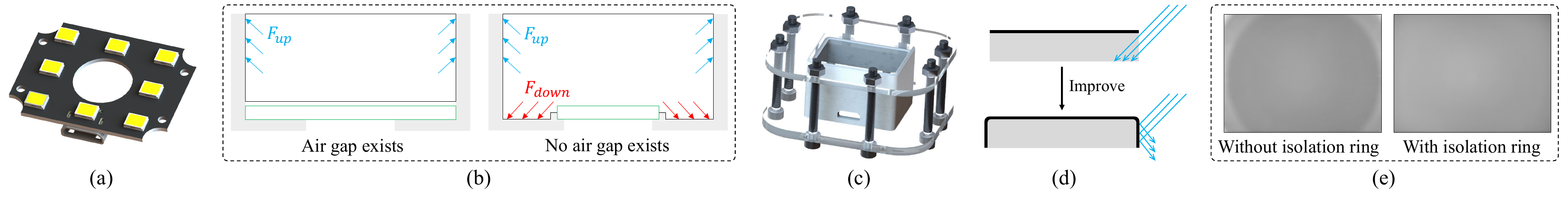}
	\caption{\revision{Design improvements of \sensor{}. (a) Eight LEDs distribute in a rectangular shape on the LED board. (b) The acrylic window is resized to avoid the generation of an air gap. (c) The mold used for fabricating the translucent gel layer is secured with eight screws. (d) The four sides of the translucent gel layer are also coated with black gel to prevent ambient light from entering. (e) Two reference images captured without and with installing the isolation ring. }}
	\label{fig:improvement}
\end{figure*}

\textbf{LED board.} In the previous version, DTact sensor~\cite{lin2023dtact}, the utilization of an LED ring may produce uneven illumination, characterized by a brighter light intensity at the center than that at the periphery. To mitigate this problem, we design a compact LED board with eight LEDs evenly arranged in a rectangular shape on it \revision{as shown in~\fig{fig:improvement}~(a)}. Furthermore, we incorporate a CN5711 integrated circuit to regulate the current inputs for the LEDs, which helps to provide stable and consistent illumination for the sensor. The LED board is secured to the sensor base by means of four locating holes and is powered by a 5$\text{V}$ USB port.

\textbf{Sensor base.} The sensor base, which is used to secure the camera and the LED board, is 3D printed with \revision{black nylon material (HP3DHR-PA12)} that has high strength and toughness. Furthermore, four M2 heat-set threaded inserts (labeled as 3 in~\fig{fig:design}~(a)) are installed in the sensor base, serving as connectors between the \sensor{} sensor and other platforms such as robot grippers and dexterous hand fingers.

\textbf{Sensor shell.} \revision{Attaching with the acrylic window that provides a clear window for the camera, the sensor shell serves as a container for the transparent gel layer. It is 3D printed with \revision{white nylon (FS3300PA)} material which exhibits superior durability. Since the white nylon material is not opaque}, the inner base layer of the sensor shell allows light to transmit from the LED board and also helps to diffuse the light. Four M2 heat-set threaded inserts~(labeled as 7 in~\fig{fig:design} (a)) are also mounted in the bottom of the sensor shell so that it can be connected to the sensor base with four M2 screws.

\textbf{Transparent gel layer.} The transparent gel layer not only facilitates the diffusion of light to a more uniform distribution, but also serves as a transitional propagation medium with optical properties similar to those of the translucent gel layer, which helps to mitigate the issue of excessive reflection that can occur when light passes through air or other media. 
Compared to ELASTOSIL\textsuperscript{\textregistered} RT 601 silicone used in~\cite{lin2023dtact}, Hongye Jie\textsuperscript{\textregistered} 9345 silicone~(mixing ratio 1:1, shore A hardness 45) is easier to remove air bubbles with a vacuum pump. Therefore, we choose it as the material of the transparent gel layer. The mixed bubble-free silicone is poured into the sensor shell until filled. Due to the inadequate levelness of the base surface of the thermostatic oven, the sensor shell is initially placed on a horizontal optical platform at a room temperature of 25 $\tccentigrade$ for 4 hours to allow the silicone to solidify. Subsequently, the sensor shell is transferred to the thermostatic oven maintained at 50 $\tccentigrade$ for a duration of 6 hours to ensure complete hardening of the silicone.

\textbf{Acrylic window.} In the DTact sensor~\cite{lin2023dtact}, the acrylic window is designed to align with the inner dimensions of the sensor shell. As a result, an air gap often exists between the transparent gel layer and the acrylic window after the transparent silicone cures\revision{, as~\fig{fig:improvement}~(b) illustrates}. This is because the rough inner walls of the sensor shell have much higher adsorption capacity than that of the smooth surface of the acrylic window. The air gap is  squeezed out when the sensor comes into contact with objects; thus the tactile image brightens overall because the light from the translucent gel layer transmits to the camera without decaying through the air gap. To eliminate this air gap, we reduce the size of the acrylic window, so that the rough inner base of the sensor shell increases the downward adsorption force to the transparent gel layer. The acrylic window is securely attached to the nested frame of the sensor shell with waterproof glue.

\textbf{Translucent gel layer.} The translucent gel layer is used to reflect light, which forms the fundamental principle of the 3D shape reconstruction function of \sensor. In the previous version as described in~\cite{lin2023dtact}, a mold is attached to the sensor shell, and the translucent silicone is poured into the mold and left to cure to the translucent gel layer. Although this method is relatively easy to fabricate and install, it has some drawbacks. For instance, the thickness of the sensor shell is increased due to the requirement of four holes to install the mold. Furthermore, the success rate of curing the translucent silicone is reduced, since only four screws can not provide sufficient force to secure the mold in place. To address these issues, in \sensor, we develop a base board and attach it to the sensor shell through the four heat-set threaded inserts. The acrylic mold for the translucent gel layer is then connected to the base board using eight pairs of M3 screws and nuts\revision{, as shown in~\fig{fig:improvement}~(c)}. The tightness resulting from this fastening mechanism ensures that the mold remains stable throughout fabrication, thus strongly guaranteeing successful outcomes. In addition, to augment the adhesive force of the translucent gel layer, its size is intentionally designed to exceed the inner dimensions of the sensor shell as~\fig{fig:design}~(b) shows. The material for the translucent gel layer is translucent Posilicone\textsuperscript{\textregistered} DRSGJ02 silicone (mixing ratio: 1: 1, shore A hardness 5). After pouring the mixed bubble-free silicone into the mold, it takes 6 hours for the silicone to cure at 25 $\tccentigrade$.

\textbf{Black gel layer.} The black gel layer serves two purposes. First, it absorbs the inner light that transmits through the translucent gel layer. Second, it prevents ambient light from entering the translucent gel layer. However, in the case of the DTact sensor~\cite{lin2023dtact}, only the upper surface of the translucent gel layer is coated with a thin black gel layer. This inevitably results in ambient light transmitting into the translucent gel layer through its four sides, leading to fluctuations in the brightness of tactile images. Therefore, we replace the mold for the translucent gel layer after it cures with one that has larger thickness and inner frame size, which ensures that the four sides of the translucent gel layer are also covered with black gel\revision{, as shown in~\fig{fig:improvement}~(d)}. Instead of using the same silicone as the translucent gel layer for the black gel layer, we opt for Smooth-On\textsuperscript{\textregistered} Ecoflex 00-30 silicone (mixing ratio: 1: 1, shore 00 hardness 30) for its superior durability. To give it a black color, we add some black silicone pigment to the mixture. With the LED board lighting from below, we apply the black silicone onto the translucent gel layer until it completely blocks the light.

\textbf{Isolation ring.} The surface of the acrylic window reflects light from the vertical surfaces of the inner base layer of the sensor shell, which can interfere with the light transmission in the peripheral areas of the translucent gel layer. This is because the light from the inner base layer is much stronger than that from the translucent gel layer. As a result, the camera loses its ability to sense changes in light from these areas of the contact surface as the left image in~\fig{fig:improvement}~(e) shows. To this end, we add the isolation ring, 3D printed with \revision{black nylon material (HP3DHR-PA12)}, to prevent such strong light from transmitting from the vertical surfaces of the inner base layer to the acrylic window. The right image in~\fig{fig:improvement}~(e) shows the corrected sensed tactile surface.

\begin{table}[t]
\centering
\caption{The bill of materials (BOM) for fabricating a \sensor{} sensor.}
\begin{tabular}{lccc}\toprule
Component & Description & Process\\
\midrule
Glue & YLG-YKL500  & \multicolumn{1}{c}{\multirow{4}{*}{Off-the-shelf}}\\
Nuts & 8 M2-3-5, 8 M3 & \\
Screws & 4 M2-6, 8 M3-30 & \\
Camera & Frank-S15-V1.0-$160^\circ$ & \\
\midrule
LED Board & $28\times21\times4 mm$ & PCB soldering\\
\midrule
Isolation Ring & \revision{Black nylon (HP3DHR-PA12)} & \multicolumn{1}{c}{\multirow{3}{*}{3D printing}}\\
Sensor Base & \revision{Black nylon (HP3DHR-PA12)} & \\
Sensor Shell & \revision{White nylon (FS3300PA)} & \\
\midrule
Acrylic Window & $2mm$ thick acrylic board & \multicolumn{1}{c}{\multirow{4}{*}{Laser cutting}}\\
Base Board & $3mm$ thick acrylic board & \\
The First Mold & $2.5mm$ thick acrylic board & \\
The Second Mold & $2.8mm$ thick acrylic board & \\
\midrule
Transparent Gel & Hongye Jie\textsuperscript{\textregistered} 9345 & \multicolumn{1}{c}{\multirow{3}{*}{\makecell{Silicone\\processing} }}\\
Translucent Gel & Posilicone\textsuperscript{\textregistered} DRSGJ02 & \\
Black Gel & Smooth-On\textsuperscript{\textregistered} Ecoflex 00-30 & \\
\bottomrule
\end{tabular}
\label{tab:components}
\end{table}

\begin{figure*}[ht]
	\centering
	\includegraphics[width= \linewidth]{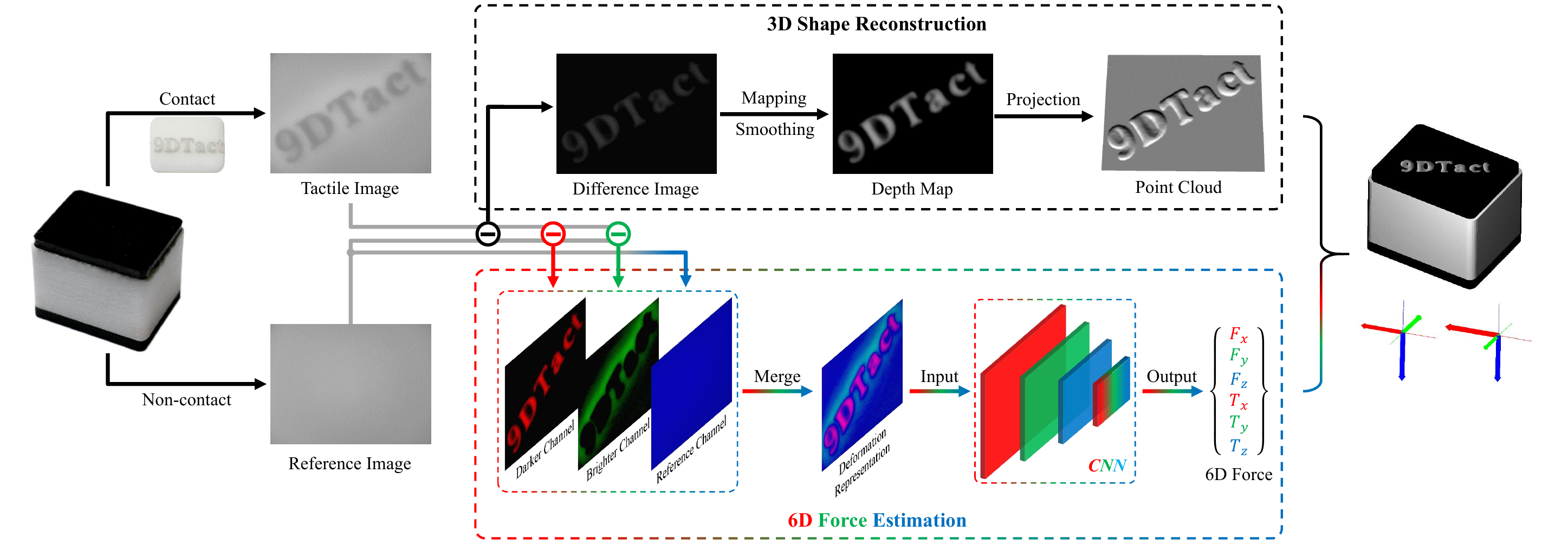}
	\caption{Pipeline for \sensor{}'s two key functions. \sensor{} utilizes a modeling-based method for 3D shape reconstruction and a learning-based method for 6D force estimation. The reference image is captured when there is no contact on \sensor{}, while the tactile image is captured when a badge with the text ``\sensor{}'' is pressed on \sensor{}. }
	\label{fig:pipeline}
\end{figure*}

\subsection{Conclusion and Comparison of the Sensor Design}
\tab{tab:sensors-comparison} compares \sensor{} with existing flattened compact vision-based sensors, while \tab{tab:components} summarizes some detailed information for each component.
Here, we summarize the outstanding characteristics of \sensor{}:
\begin{itemize}
    \item {\textbf{Compact.} The \sensor{} sensor is remarkably compact, with the size of only $32.5 \text{mm} \times 25.5 \text{mm}\times 25.5 \text{mm}$, which is approximately $22\%$ the size of the DTact sensor~\cite{lin2023dtact}. With the smallest volume among the sensors listed in~\tab{tab:sensors-comparison}, \sensor{} is adaptable to be installed in a wide range of robotic platforms, from grippers to dexterous hands. Moreover, \sensor{} features a relatively large sensing area, resulting in the smallest dimension-to-sensing area ($D/A$) ratio of all the sensors in~\tab{tab:sensors-comparison}.}

    \item {\textbf{Robust.} Our improvement in the black gel layer makes \sensor{} robust to dynamic ambient light. Furthermore, in Section ~\ref{sub:force_data}, we press the objects with sharp geometries against a single \sensor{} sensor to collect over $100,000$ images. Remarkably, the sensor factors, such as imaging and illumination, remain stable throughout the experiment, and the contact surface shows no visible signs of damage.}
    
    \item {\textbf{\revision{Easy-to-fabricate} and low-cost.} As summarized in~\tab{tab:components}, the components are fabricated with minimal effort, utilizing highly commercialized conventional machining processes such as 3D printing, laser cutting, and printed circuit board (PCB) soldering. Besides, each component is subject to strict mechanical positional constraints during assembly, thereby minimizing the need for additional adjustments and reducing performance differences in sensors caused by variations in assembly. To further support the replication of our \sensor{} sensor, we also provide a step-by-step video tutorial that details the entire fabrication and assembly process, enabling researchers with varying levels of experience to easily reproduce the sensor on their own. Furthermore, building a \sensor{} only costs about $\$15$, which includes two reusable molds.}

\end{itemize}


\section{3D Shape Reconstruction}
\label{sec:shape_reconstruction}
	\begin{figure}[t]
	\centering
	\includegraphics[width= \linewidth]{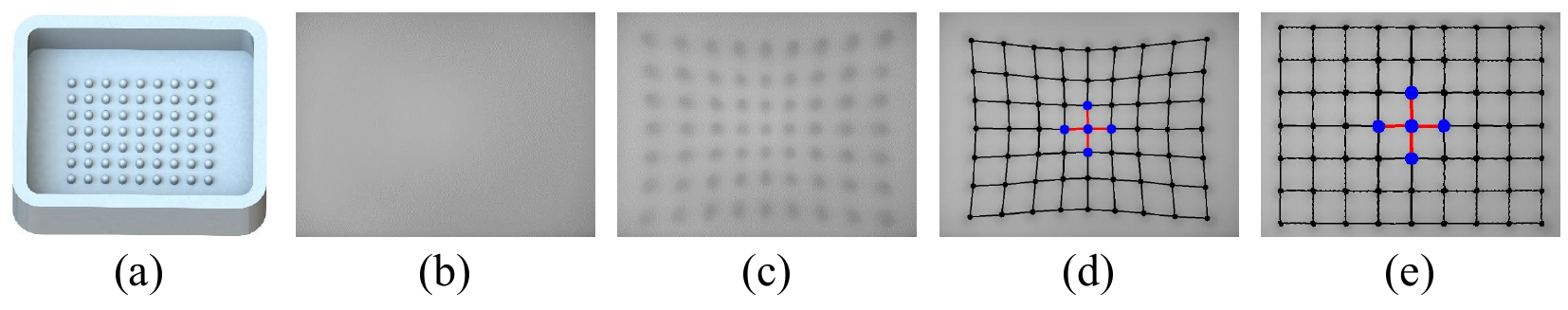}
	\caption{Camera calibration. (a) The calibration board features a cylinder array. (b) The reference image. (c) The tactile image is captured when pressing the calibration board on \sensor{}. (d) Imprints from the calibration board are detected. (e) The result of rectifying and cropping the image in (d).
 }
	\label{fig:calibration}
\end{figure}

\subsection{\revision{Reconstruction Pipeline}}
\revision{\sensor{} inherits the 3D shape reconstruction method of DTact~\cite{lin2023dtact}, the pipeline of which is illustrated in~\fig{fig:pipeline}. \cd{Both the tactile and reference images are converted to grayscale images because the reconstruction process only relies on pixel luminance. The difference image, calculated by subtracting the reference image from the tactile image, is mapped to a depth map with a calibrated mapping list. This mapping list is calibrated using the ``single image'' calibration approach proposed in~\cite{lin2023dtact}, which is efficient for requiring only a single image for calibration.} In addition, we apply two continuous Gaussian filters to denoise the depth map. Finally, the depth map is converted to point clouds to render and visualize the sensor surface.}

\begin{figure}[t]
	\centering
	\includegraphics[width= \linewidth]{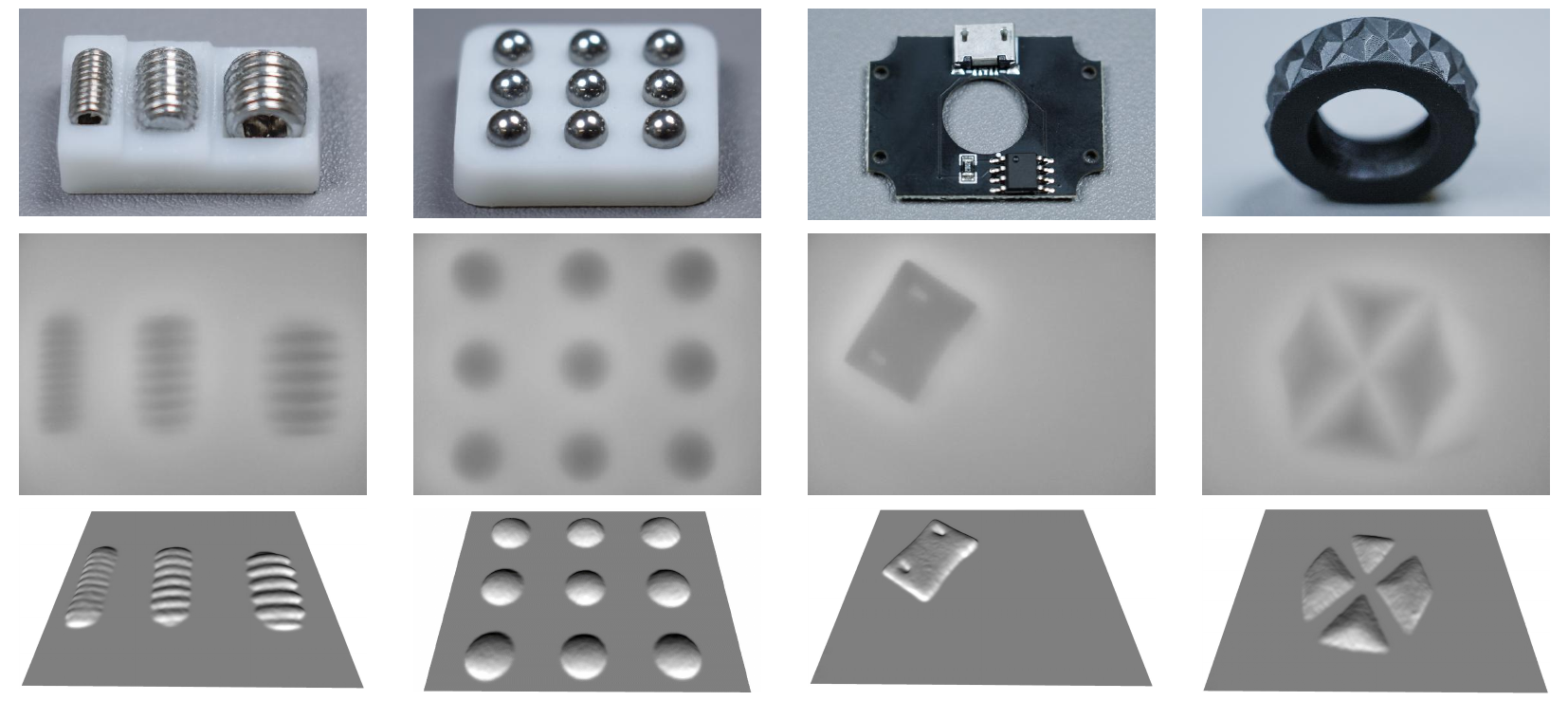}
	\caption{3D shape reconstruction results of three set screws (M4, M6, and M8), a ball array, a micro USB port, and a wheel hub model. They are visual images, tactile images, and reconstructed point clouds from top row to bottom.}
	\label{fig:reconstruction}
\end{figure}

\subsection{Camera Calibration}
In order to obtain accurate tactile images, we need to rectify the optical distortion introduced by the wide-angle camera and the three gel layers. The marker-based image rectification method introduced by GelSlim3.0~\cite{taylor2022gelslim} has the potential to correct distortion caused by multiple factors. However, markers are not painted on \sensor{} for the reasons described in~\ref{sec:related_force}. Therefore, a calibration board, as shown in~\fig{fig:calibration}~(a), is designed with a cylinder array and a frame that fits into the sensor shell of \sensor{}. When the calibration board is pressed on the sensor surface, the cylinders imprint a grid array of virtual markers as shown in~\fig{fig:calibration}~(c). \fig{fig:calibration}~(d) shows that markers are detected using algorithms from OpenCV~\cite{bradski2000opencv}. 

In contrast to GelSlim3.0~\cite{taylor2022gelslim} that regards the outermost points as anchor points, we choose the five relatively central points in blue because of their minimal aberration. The rectified positions of other points in the image frame can be computed by extending the five anchor points to equidistant grids. With the detected positions and rectified positions of all markers, we can compute the mapping array to rectify image. Finally, we crop the rectified image from $640 \times 480$ resolution to $460 \times 345$ as shown in~\fig{fig:calibration}~(e). In summary, the virtual markers-based image rectification method employed by \sensor{} enables simultaneous calibration of these parameters:
\begin{enumerate}
    \item {The lens distortion; }
    \item {The pixel position on the tactile image that corresponds to the sensor surface's central position;}
    \item {The physical length on the sensor surface corresponding to one pixel of the tactile image.}
\end{enumerate}

\subsection{Improvements of 3D Shape Reconstruction}
\label{sub:shape_reconstruction}

\revision{Although DTact~\cite{lin2023dtact} demonstrates excellent performance in shape reconstruction,} it is limited to reconstructing only the central area of its surface due to uneven illumination and significant disturbance from environmental light in the peripheral area, as discussed in~\ref{sec:design_details}. \revision{In contrast, \sensor{}, through its refined design and advanced calibration technique, achieves comprehensive surface reconstruction with an incremental yet significant improvement in precision.}

\cd{To validate its performance, \sensor{} employs the same approach as proposed in~\cite{lin2023dtact}. This involves pressing a metal ball, different in radius from that used in the calibration phase, at various positions on the sensor surface to capture 20 distinct images. These images are then processed to compute actual depth maps, using circle detection algorithms in OpenCV~\cite{bradski2000opencv}. The quantitative results of \sensor{}'s reconstruction precision yield a mean absolute error (MAE) of $0.0462 \text{mm}$ and a standard deviation (Std) of $0.0304 \text{mm}$. While DTact records an MAE of $0.0476 \text{mm}$ and an Std of $0.0352 \text{mm}$.}

\fig{fig:reconstruction} showcases several reconstruction examples, highlighting the intricate geometric details captured by \sensor{}.


\section{6D Force Estimation}
\label{sec:force_estimation}

	\begin{figure}[t]
	\centering
	\includegraphics[width=\linewidth]{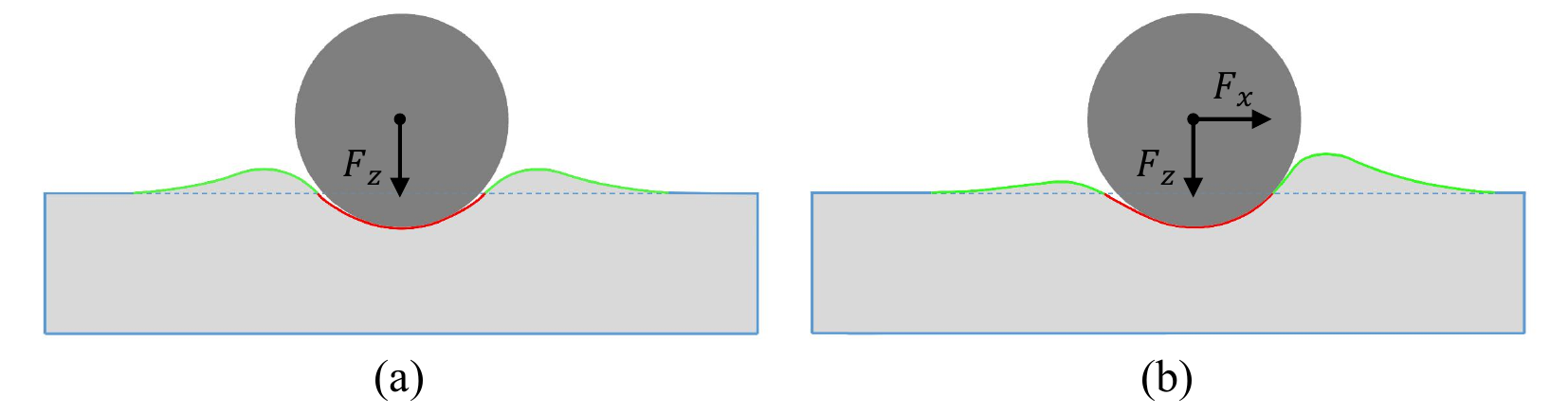}
	\caption{Deformable property of the translucent gel. The dotted line represents the original thickness of the gel. The areas that are thinner than the original thickness are colored in red, while the thicker areas are in green. (a) The object is pressed on the gel. (b) The object is dragged to the right after being pressed.}
	\label{fig:gelflow}
\end{figure}

\subsection{Dense Deformation Representation}
\label{sub:gel_flow}

As mentioned in~\ref{sec:shape_reconstruction}, the pixels within the contact areas become darker and are utilized to compute the 3D contact geometry. Surprisingly, we also observe a concurrent increase in the luminosity of pixels surrounding the contact areas. To elucidate this phenomenon, it is necessary to consider the flow properties of the gel.

\begin{figure}[t]
	\centering
	\includegraphics[width= \linewidth]{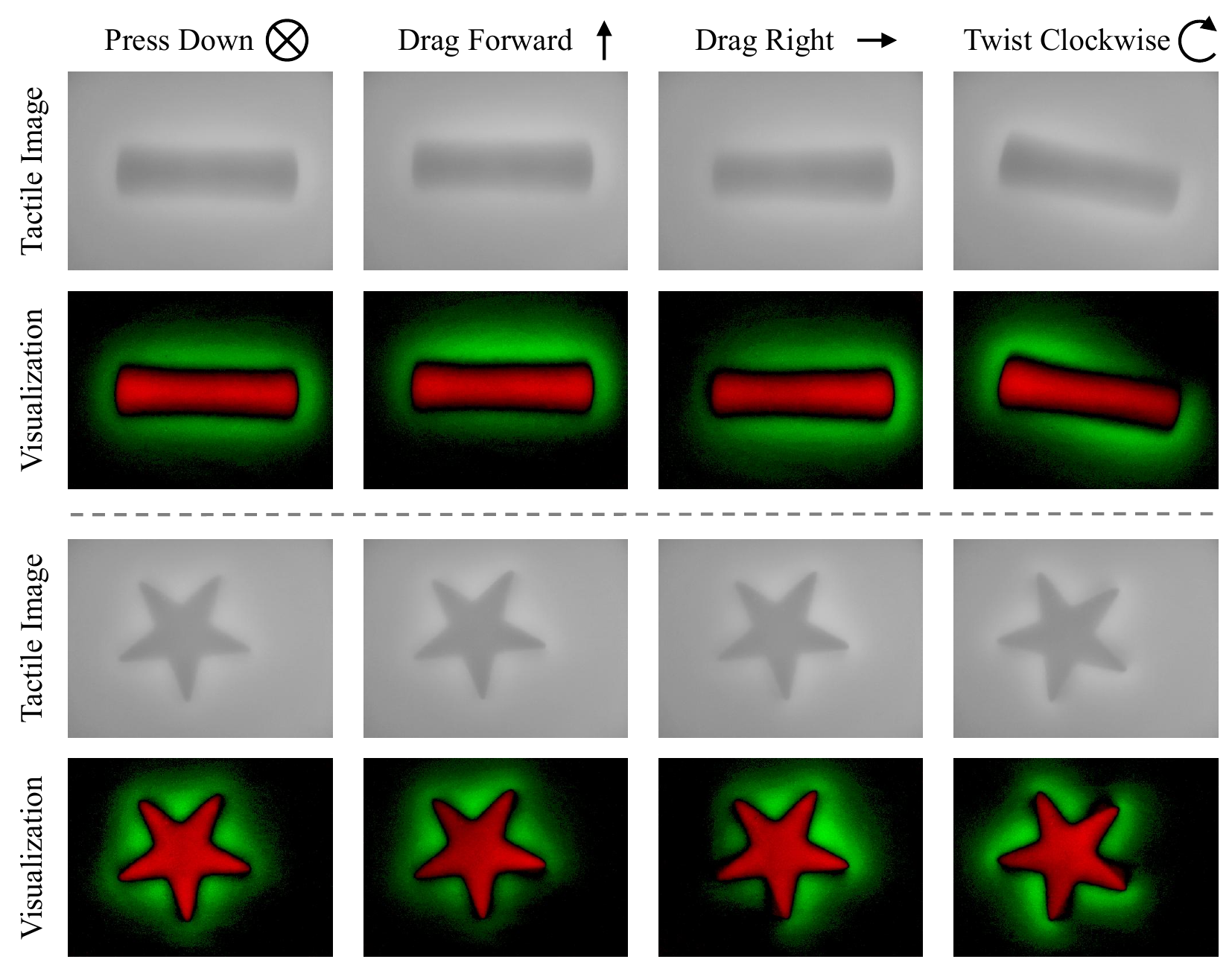}
	\caption{A cylinder and a star-shape object contact with \sensor{} respectively. The object is firstly pressed onto \sensor{}, and then dragged forward, dragged to the right, and twisted clockwise while maintaining contact. The tactile images are captured by the camera, and they are visualized as the visualization images with the darker areas in red and the brighter areas in green.}
	\label{fig:gelflow-representation}
\end{figure}

As a hyper-elastic material, the pressed gel tends to flow outward to its neighboring regions, resulting in an increase in thickness in the surrounding areas, as shown in~\fig{fig:gelflow}~(a). Similarly, when the contact object applies shear force or twist force, the surrounding gel will flow to accumulate along the moving direction, as shown in~\fig{fig:gelflow}~(b). Furthermore, according to the principle that thinner contact areas result in darker pixels, the thicker surrounding regions induce brighter pixels. Eventually, as the visualization images illustrated in~\fig{fig:gelflow-representation} with darker areas colored red and brighter areas green, the green pixels turn to be much brighter in the object's moving direction. Therefore, with the darker areas extracting the concave deformation information and the brighter areas extracting both the convex deformation and shear deformation information,  deformation in all directions can be encoded in such dense deformation representation as the visualization images in~\fig{fig:gelflow-representation} show.

Our goal is to reconstruct the 6D force from the described dense deformation representation. It is unlikely to use numerical methods such as iFEM because the physical thickness of the bulging areas can not be acquired. Therefore, we leverage deep convolutional neural networks, which have shown great potential in image information extraction~\cite{krizhevsky2012imagenet}. Following, we will introduce physical configurations to collect dataset, details on model training, and results of 6D force estimation.

\begin{table*}[ht]
\centering
\caption{Configuration and hyper-parameters for force estimation deep model training.}
\begin{tabular}{cccccccc}\toprule
Model & Batch size & Learning rate & Pretrained & Optimizer & Loss Function 
& Total epoch & GPU \\
\midrule
Densenet-169& 64 & $5 \times 10^{-4}$ & No & Adam & MAE (Sum) & 200 & Nvidia A40\\
\bottomrule
\end{tabular}
\label{tab:model_training}
\end{table*}

\begin{table*}[ht]
\centering
\caption{Validation results of 6D force estimation on two test sets (force in $N$, torque in $Nm$).}
\setlength{\tabcolsep}{5pt}
\begin{tabular}{cccccc}\toprule
Splitting Method & Training set & Test set & Selected epoch & Mean absolute error (MAE) & Standard deviation (Std)\\
\midrule
\revision{Standard} & \revision{90417} & \revision{10000} & \revision{153} & \revision{[0.30, 0.35, 0.28, 0.009, 0.008, 0.001]} & \revision{[0.26, 0.30, 0.32, 0.009, 0.008, 0.001]}\\
\revision{Object-based} & \revision{89995} & \revision{10422} & \revision{189} & \revision{[0.35, 0.40, 0.41, 0.011, 0.010,  0.002]} & \revision{[0.31, 0.36, 0.44, 0.015, 0.014, 0.003]}\\
\bottomrule
\end{tabular}
\label{tab:results}
\end{table*}

\subsection{Data Collection and Splitting}
\label{sub:force_data}
Previous tactile image datasets~\cite{do2023densetact, gomes2021generation, church2022tactile, chen2022bidirectional} are mainly collected by mounting several objects on an autonomous machine, resulting in limited flexibility due to the inconvenience of object swapping, and thus, severely restricting the diversity of contact geometry. Furthermore, the process of force label collection dose not require recording the pose of the contact object. Therefore, we opt to manually press objects to increase the diversity of objects and the flexibility of pressing and swapping them.

\begin{figure}[t]
	\centering
	\includegraphics[width= \linewidth]{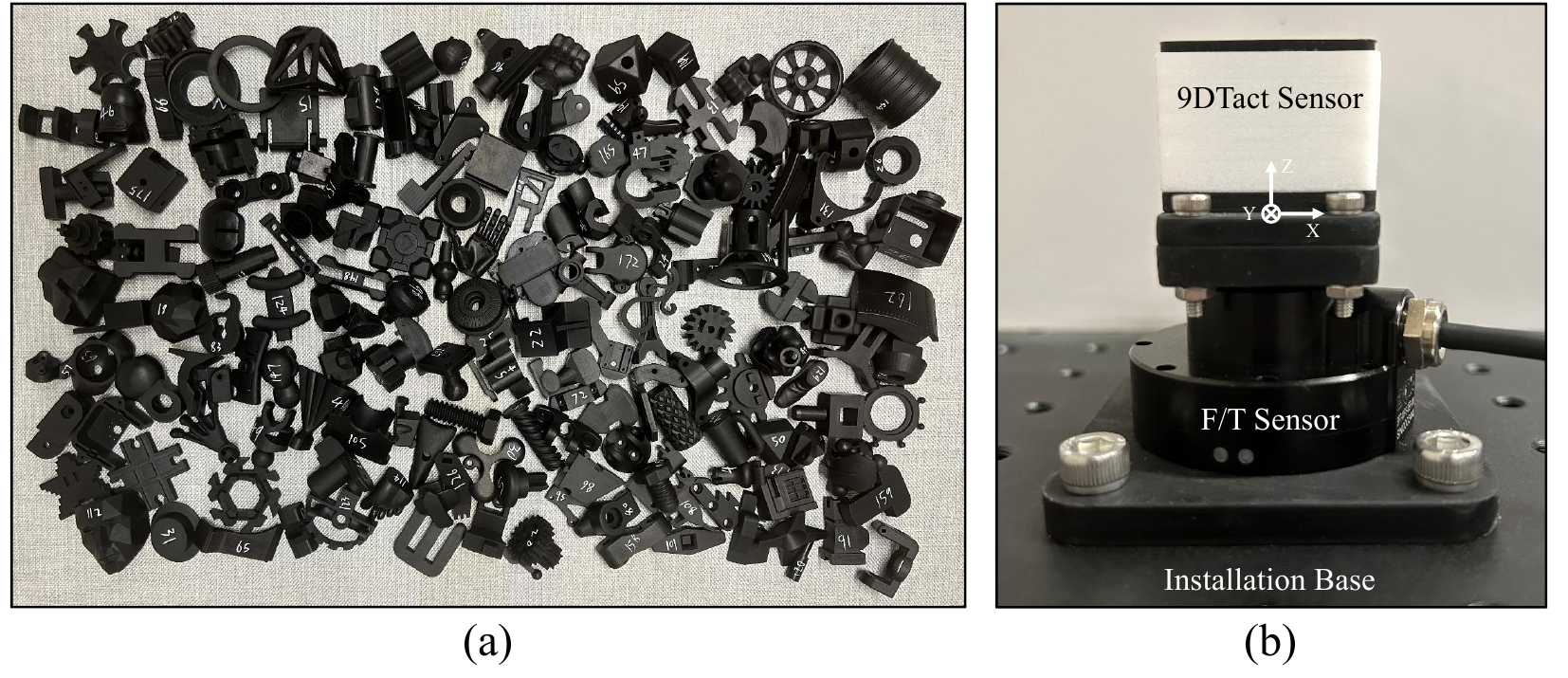}
	\caption{Physical configurations for dataset collection. (a) 175 objects with various geometric shapes are 3D-printed for collecting data. (b) The \sensor{} sensor is installed on a BOTA MiniONE Pro 6-axis F/T sensor to collect 6D force labels.}
	\label{fig:setup}
\end{figure}

We handpick 175 CAD models with various geometric shapes from the Thingi10K~\cite{zhou2016thingi10k} dataset, and 3d-print them with black resin material, as shown in~\fig{fig:setup}~(a). For the hardware setup, as~\fig{fig:setup}~(b) shows, \sensor{} is fastened on a BOTA MiniONE Pro 6-axis F/T sensor which provides precise 6D force labels. To efficiently collect data, we develop a program to autonomously sample image-force pairs so that we only need to press the objects on \sensor{} in \revision{different} orientations evenly. More specifically, a data pair is saved only when contact is detected and the 6d force is significantly different from all the forces saved during the same continuous contact period.

As introduced in~\sec{sub:gel_flow}, \sensor{} is able to extract pressing, dragging, and twisting motions of the contact geometry. Therefore, we not only press the objects but also drag and twist them on \sensor{} to collect data with various contact status. Finally, we collect 100,417 image-force pairs totally from a single \sensor{} sensor, taking about 10 hours cumulatively.

Leaving the \revision{remaining} data as the training set, the test set is selected based on two splitting methods.
\begin{itemize}
    \item {\textbf{\revision{Standard} splitting.} \revision{10000} pairs of data are randomly selected from all data as test set. This splitting strategy is generally used to test the \revision{model's in-distribution generalization capability}.}
    \item {\textbf{Object-based splitting.} \revision{18} of the 175 objects are randomly selected as test objects, and all data sampling from these objects serves as test set, which consists of \revision{10422} pairs of data in our case. This splitting strategy aims to test the model's ability to generalize to unseen objects.}
\end{itemize}

\subsection{Details of the Neural Network}
\begin{figure}[t]
	\centering
	\includegraphics[width= \linewidth]{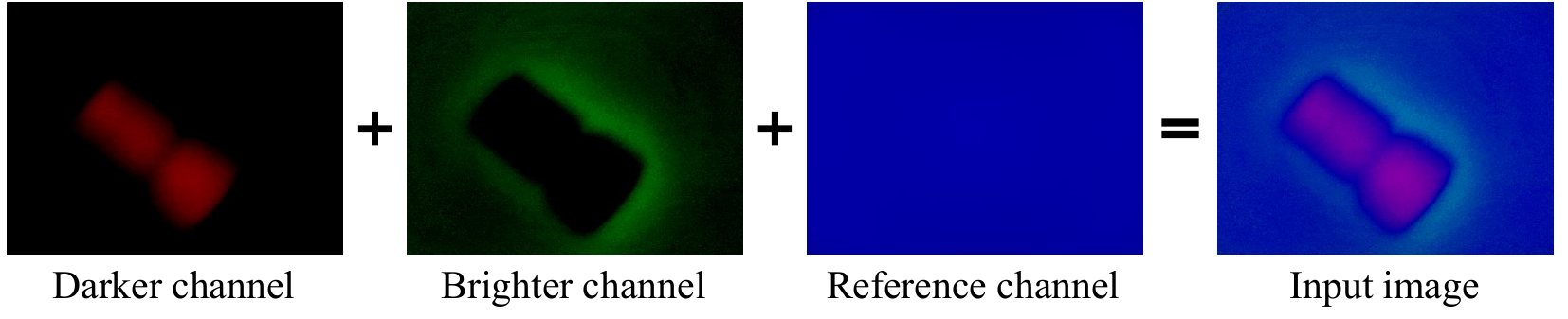}
	\caption{Input image for predicting 6D force. It consists of three channels: the darker channel, the brighter channel, and the reference channel.}
	\label{fig:input_image}
\end{figure}

\textbf{Input images.}
In order to provide sufficient and well-defined physical information for the neural network, the input image's three channels are replaced by the darker image, the brighter image, and the grayscale reference image, as shown in~\fig{fig:input_image}. The darker image, which contains information about the contact geometry, is generated by subtracting the tactile image from the reference image. It is the same as the difference image used for shape reconstruction as illustrated in~\fig{fig:pipeline}. The brighter image reveals information about the dragging and twisting motions of the contact geometry, and it is obtained in a similar way as the darker image, but with the objects of subtraction reversed. The reference image is captured each time the sensor is initiated, which differs slightly for different continuous usage periods and thus provides information of the sensor's initial state.

\textbf{Neural Networks.}
We select Densenet~\cite{huang2017densely} as the neural network for predicting the 6D force. We utilize the implementation of Densenet from PyTorch library~\cite{paszke2019pytorch}, and modify the output channel of the fully connected layer to $6$. Two models are trained on datasets split using the two splitting methods, with the same training configuration depicted in~\tab{tab:model_training}. We also train Resnet~\cite{he2016deep} and ViT~\cite{dosovitskiy2020image}, but they perform worse than Densenet. Therefore, we only present the details and following results of Densenet, as our focus in this paper is to validate that our proposed dense deformation representation is able to provide comprehensive deformation information.

\begin{figure*}[t]
\centering
      \includegraphics[width= \linewidth]{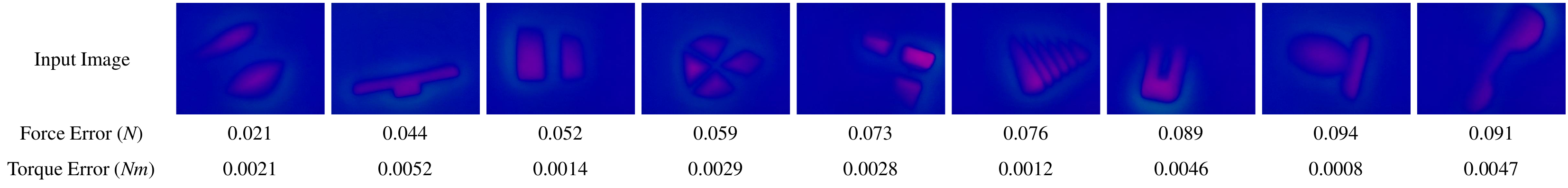}
\caption{Results from the test set, using the standard splitting method, highlight our dataset's complexity in both geometry and motion dynamics. The average force and torque errors demonstrate the generalization capability of the trained deep model.
}
\label{fig:estimation_examples}
\end{figure*}

\subsection{Results of 6D Force Estimation}

As the quantitative validation results in~\tab{tab:results} shows, the models are able to \revision{estimate} accurate 6D force and generalize to both unseen contact status and objects. For \revision{standard} validation, the absolute mean errors are $\revision{0.307} N$ for forces and $\revision{0.006} Nm$ for torques. For object-based validation, the errors are $\revision{0.370} N$ and $\revision{0.0077} Nm$ respectively.
\cd{While these quantitative metrics are informative, direct comparisons of \sensor{} with other sensors remains complex for several reasons. First, in contrast to other methods listed in~\tab{tab:Force-comparision} that validate their methods with a small set of simple objects, our test set comprises images with intricate geometries and in-plane motions as shown in~\fig{fig:estimation_examples}.
Second, different studies report varied metrics for normal and tangential force estimation such as an RMSE of \text{$0.62N$} in~\cite{yuan2017gelsight} and an MAE of \text{$0.41N$} in~\cite{do2023densetact}.  
Lastly, inherent challenges in hardware comparisons arise due to factors such as resource limitations for sensor replication and performance variations attributed to the fabrication process.}

\cd{Drawing from the comprehensive analysis of methodology, dataset, and results,} our method for 6D force estimation exhibits superiority in physical simplicity due to the integration of the optical and deformable properties of the translucent gel, in accuracy with the help of the dense information included in the proposed deformation representation, and in generalization capability by learning from the large and resourceful dataset.


\section{Conclusion}
\label{sec:conclusion}

	In this work, we present \sensor{}, a general vision-based tactile sensor capable of 3D shape reconstruction and 6D force estimation. Specifically, we meticulously select and design each component of \sensor{} to make it compact for installation, robust for illumination,  durable for long-term usage, and simple for reproduction. We also improve 3D shape reconstruction to have a larger field ratio for reconstruction, more effective calibration procedures, and higher accuracy. Furthermore, unlike conventional methods for force estimation using painted markers, we extract a dense deformation representation from the raw tactile image by integrating the optical and deformable properties of the translucent gel. Finally, we train a force estimation neural network on a large dataset sampling from various objects with complex geometry. Empirical results show that it not only can estimate the 6D force accurately, but also can generalize to unseen geometries and objects.

\footnotesize{
\bibliographystyle{IEEEtran}
\bibliography{paper}
}

\end{document}